\documentclass{article}
\usepackage{spconf,amsmath,graphicx}

\usepackage{times}  
\usepackage{helvet} 
\usepackage{courier}  
\usepackage[hyphens]{url}  
\usepackage{threeparttable}
\usepackage{bbding}
\usepackage {graphicx,color,multirow}
\usepackage[hang,center]{subfigure}
\usepackage{booktabs}

\usepackage{caption} 
\usepackage{amssymb}
\usepackage{multirow}
\usepackage{rotating}
\usepackage{booktabs}
\usepackage{overpic}

\def\ie{\emph{i.e.}}
\def\eg{\emph{e.g.}}

\newcommand{\figref}[1]{Fig.~\ref{#1}}
\newcommand{\tabref}[1]{Tab.~\ref{#1}}
\newcommand{\secref}[1]{$\S$~\ref{#1}}
\newcommand{\eqnref}[1]{Equ.~\ref{#1}}

\definecolor{mygray}{gray}{.92}

\newcommand{\jgp}[1]{{\textcolor{red}}}

\def\ourmodel{\textit{GTNet}}

\newcommand{\mySection}[1]{\vspace{.05in}\noindent\textbf{#1~}}

\usepackage[pagebackref=true,breaklinks=true,letterpaper=true,colorlinks,bookmarks=false]{hyperref}

\title{Guidance and Teaching Network for Video Salient Object Detection}

\name{Yingxia Jiao$^{1}$, Xiao Wang$^{2}$, Yu-Cheng Chou$^{1}$, Shouyuan Yang$^{2}$, Ge-Peng Ji$^{1}$, Rong Zhu$^{1,\dagger}$, Ge Gao$^{1}$}

\address{
$^{1}$ School of Computer Science, Wuhan University
$^{2}$ Jiangxi University of Finance and Economics
}

\begin{document}
%
\maketitle
\begin{abstract}
Owing to the difficulties of mining spatial-temporal cues, the existing approaches for video salient object detection (VSOD) are limited in understanding complex and noisy scenarios, and often fail in inferring prominent objects.
To alleviate such shortcomings, we propose a simple yet efficient architecture, termed \textit{\textbf{G}uidance and \textbf{T}eaching \textbf{Net}work} (\textbf{\ourmodel}), to independently distill effective spatial and temporal cues with \textbf{\textit{implicit guidance and explicit teaching}} at feature- and decision-level, respectively.
To be specific, we 
(a) introduce a temporal modulator to implicitly bridge features from motion into appearance branch, which is capable of fusing cross-modal features collaboratively, and
(b) utilise motion-guided mask to propagate the explicit cues during the feature aggregation.
This novel learning strategy achieves satisfactory results via decoupling the complex spatial-temporal cues and mapping informative cues across different modalities.
Extensive experiments on three challenging benchmarks show that the proposed method can run at $\sim$28 \textit{fps} on a single TITAN Xp GPU and perform competitively against 14 cutting-edge baselines.
\end{abstract}

\section{Introduction}\label{sec:intro}
Video salient object detection (VSOD) has been a longstanding research topic in the area of computer vision, which aims to predict conspicuous and attractive objects in a given video clip.
It has been applied to autonomous cars, action segmentation, and video captioning.
In recent years, much progresses~\cite{Fan2019ShiftingMA,Feichtenhofer2019SlowFastNF} have been witnessed in handling unconstrained videos, but there remains a large room to improve that has not yet been adequately explored.

Motion (\eg, optical flow~\cite{Teed2020RAFTRA} and trajectory~\cite{Keuper2015MotionTS}) and appearance features (\eg, color~\cite{Wei2012GeodesicSU} and superpixel~\cite{Wang2015SaliencyawareGV} segmentation) are both crucial cues for understanding the dynamic salient object with an unconstrained background.
Several works have been made to develop spatial-temporal convolution neural networks (CNNs) for learning discriminative appearance and motion features, in which recurrent memory~\cite{Li2018FlowGR,Song2018PyramidDD} and 3D convolution~\cite{ji2021pnsnet} are frequently used.
However, they are hindered by the following problems.
For the former, it is unable to handle spatial and temporal cues simultaneously. 
Besides, it only processes the inputs sequentially due to the transmissible temporal memory, and thus, its efficiency is largely limited.
As for the latter, 3D CNNs are difficult to optimise if the number of temporal layers is large, due to exponential growth of the numerical solution space.
In addition, it is overloaded by high computation cost ($\sim$1.5$\times$ memory cost than 2D CNNs).
Thus, it is imperative to separately model a spatial and temporal representation scheme, which can bring considerable benefits for VSOD.

\begin{figure}
    \centering
    \begin{overpic}[width=\linewidth]{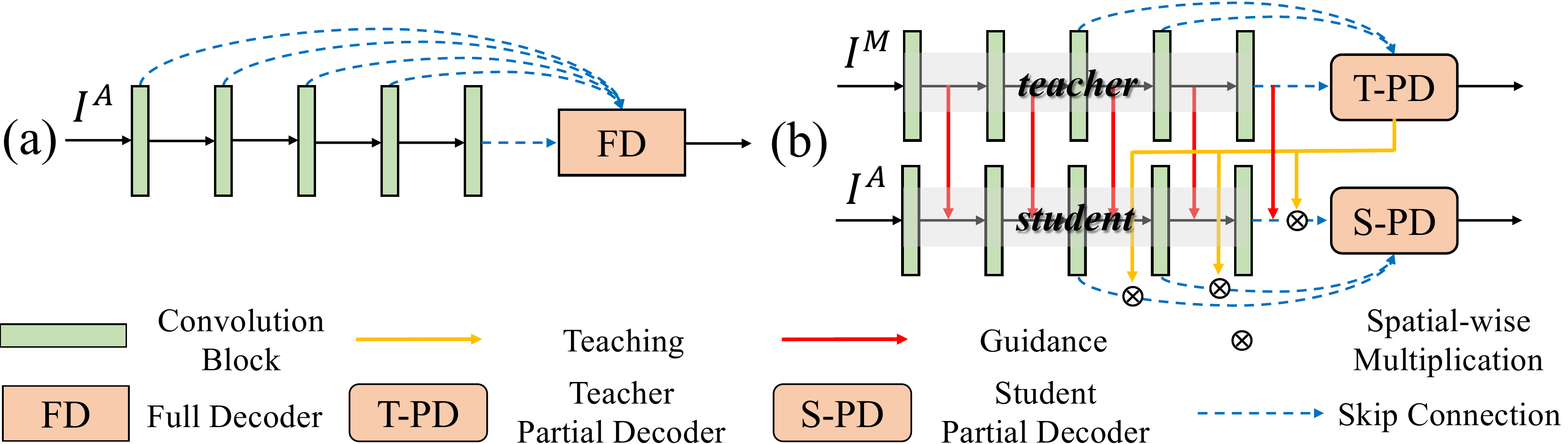}
    \end{overpic}
    \caption{Illustration of the proposed \textit{guidance and teaching strategies}.
    (a) typical UNet-shaped~\cite{ronneberger2015u} framework with full decoder for aggregating feature pyramids.
    (b) our pipeline utilise implicit guidance to bridge teacher (\ie, motion-dominated) branch and student (\ie, appearance-dominated) branch.
    For requiring explicit knowledge from the teacher branch, we utilise the teacher partial decoder (T-PD) under deep supervision to get the motion-guided mask, and use it to teach the decoding phase of student partial decoder (S-PD).
    }
\label{fig:aggregation_strategy}
\end{figure}

\begin{figure*}[t!]
    \centering
    \begin{overpic}[width=.89\linewidth]{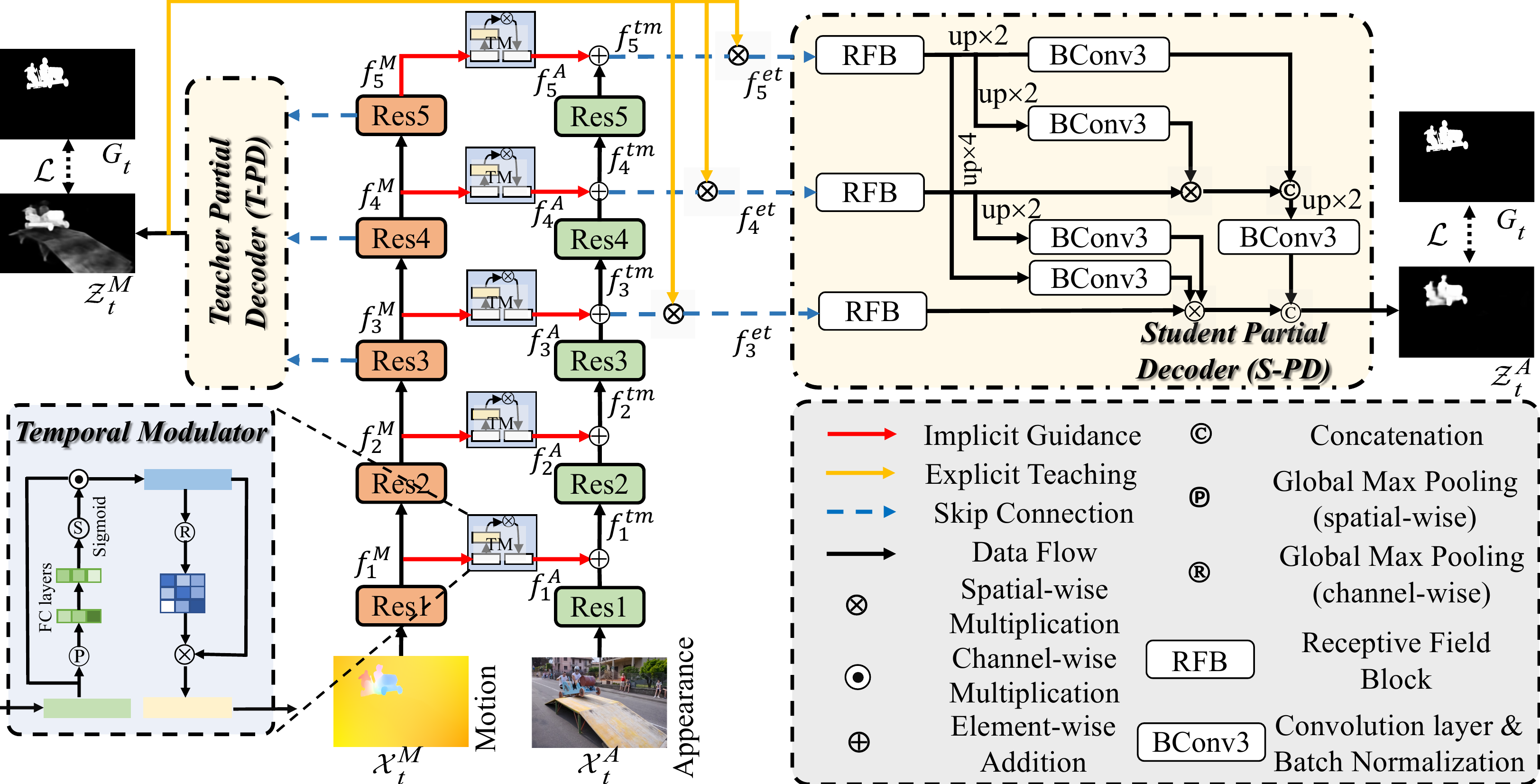}
    \end{overpic}
    \caption{The illustration of our~\ourmodel. The dual-branch framework is bridged by the \textit{implicit guidance} (red lines) in the temporal modulator.
    The student partial decoder excavates the appearance-dominated features under the \textit{explicit teaching} (yellow lines) from the motion-guided mask, which stems from the teacher partial decoder.
    Please refer to~\secref{sec:method} for more details.}
\label{fig:framework}
\end{figure*}

To efficiently fuse the motion and appearance cues, research has explored different cross-modal strategies with separate branches, and has achieved encouraging results.
Earlier works (\eg, \cite{jain2017fusionseg}) address this issue via encoding motion and appearance features individually and directly aggregating them.
However, it may unintentionally introduce feature conflicts due to different characteristics between them.
The other, more natural idea is to excavate the relationship between motion and appearance cues in a guided fashion.
Thus, recent cutting-edge methods (\eg,\cite{Li2019MotionGA}) encourage utilising the motion-dominated features to guide the encoding/decoding of appearance-dominated features, achieving satisfactory results.
However, these methods serve it as the problem of implicit feature propagation, which is primarily hindered by its inexplicable feature transmission/learning process.

To alleviate the above concerns, we propose a framework, equipped with \textit{implicit guidance and explicit teaching strategies} at feature- and decision-level, towards effective and efficient VSOD.
As shown in~\figref{fig:aggregation_strategy}, the proposed scheme serves the motion-dominated feature as teacher knowledge. It then uses these features to implicitly guide the encoding of appearance-dominated (\ie, student) features, while explicitly teaching the decoding of student partial decoder via the teacher partial decoder (T-PD).
Attributing to these strategies, this efficient scheme can generate satisfactory results in challenging scenarios.

To the best of our knowledge, \ourmodel~is the pioneering work to explore both implicit guidance and explicit teaching mechanism for VSOD.
Our main contributions are as follows:
\vspace{-10pt}
\begin{itemize}

\item We emphasise the importance of both implicit guidance and explicit teaching strategies for the spatial-temporal representations. It is based on the observation that motion-guided features and masks provide discriminative semantic and temporal cues without redundant structures, contributing to efficient decoding phase of an appearance-dominated branch.

\vspace{-5pt}
\item We introduce the temporal modulator that contains two sequential attention mechanisms from channel and spatial view, working in a deeply collaborative manner.

\vspace{-5pt}
\item Comprehensive experiments on 3 benchmarks demonstrate that the proposed method can run at $\sim$28 \textit{fps} on a single TITAN Xp GPU and get competitive performance against 14 cutting-edge VSOD models using 3 metrics, making it a potential solution for practical application.

\end{itemize}


\vspace{-5pt}
\section{Method}\label{sec:method}
\vspace{-5pt}
\subsection{Overview of Framework}
\vspace{-5pt}
Given a series of input frames $\{ \mathcal{X}^A_t \}_{t=2}^{T}$ and corresponding optical flow map $\{ \mathcal{X}^M_t \}_{t=2}^{T}$, which is generated by optical flow generator (\ie, RAFT~\cite{Teed2020RAFTRA}).
Note that we discard the first frame $\mathcal{X}^A_1$ and optical flow map $\mathcal{X}^M_1$ in our experiment due to the impact of the frame-difference algorithm.
As shown in~\figref{fig:framework}, we first feed $\mathcal{X}^A_t$ and $\mathcal{X}^M_t$ into the proposed dual-branch architecture and generate appearance-dominated features $\{f_k^A \}_{k=1}^5$ and motion-dominated features $\{ f_k^M \}_{k=1}^5$ at frame $t$, which is implemented by two separate ResNet50 backbones.
Second, we use temporal modulator (TM, see~\secref{sec:TM}) to enhance the motion-dominated (\ie, teacher) features from  the spatial- and channel-aware view and transfer them to the appearance-dominated (\ie, student) branch  with the implicit guidance strategy.
%
%
Then, we aggregate top-three motion-dominated features $\{ f_k^M \}_{k=3}^5$ via a teacher partial decoder (T-PD) and generate a motion-guided mask $\mathcal{Z}^{M}$ at frame $t$.
This mask is used for explicitly teaching the aggregating of top-three appearance-dominated features $\{ f_k^A \}_{k=3}^5$ via another symmetric student partial decoder (S-PD) (see~\secref{sec:PD}) and generate the final prediction map $\mathcal{Z}^{A}$ at frame $t$.
%


\vspace{-10pt}
\subsection{Implicit Guidance}\label{sec:TM}
\vspace{-5pt}
To maintain the semantic consistency between these two different features, we utilise the temporal modulator (TM) for implicitly transferring the motion-dominated features from teacher branch to student (\ie, appearance-dominated) branch.
Specifically, the implicit guidance strategy works collaboratively at each feature pyramid level $k$. The function of implicit guidance strategy is formulated as:
\begin{equation}\label{eq:f^tm}
f_k^{tm} = f_k^{A} \oplus \mathcal{F}_{TM}(f_k^{M}),
\end{equation}
where $k \in \{ 1,2,3,4,5\}$. The temporal modulator function $\mathcal{F}_{TM}$ is defined as two sequential attention processes, including the channel-wise $\mathcal{A}^k_{C}$ and spatial-wise $\mathcal{A}^k_S$ attention functions at level $k$. Thus, this operation can be defined as:
\begin{equation}
\mathcal{F}_{TM}(f_k^M) = \mathcal{A}^k_S(\mathcal{A}^k_{C}(f_k^M;\mathbf{W}^k_{C});\mathbf{W}^k_{S}).
\end{equation}
Specifically, for the former one:
\begin{equation}
    \mathcal{A}^k_{C}(x;\mathbf{W}^k_{C}) = \sigma[g \langle \mathcal{P}_{max}(x);\mathbf{W}^k_{C} \rangle ] \odot x,
\end{equation}
where $\mathcal{P}_{max}(\cdot)$ denotes the adaptive max-pooling operation for the candidate feature at the spatial-aware. 
$g \langle \cdot ;\mathbf{W}^k_{C}\rangle$ means the dual fully-connected layers, which is parameterized by learnable weights $\mathbf{W}^k_{C}$. 
Besides, $\sigma[x]$ and $\odot$ represents the activation function and channel-wise multiplication operation.
In our default implementation, we adopt the widely used \textit{sigmoid} function to activate the input feature, which is formulated as $\sigma[x]=1/(1+\text{exp}(-x))$.
For the latter one:
\begin{equation}
    \mathcal{A}^k_S(x;\mathbf{W}^k_{S}) = g \langle \mathcal{R}_{max}(x) ; \mathbf{W}^k_{S}\rangle \otimes x,
\end{equation}
where $\mathcal{R}_{max}(\cdot)$ denotes the global max-pooling operation along the channel dimension for the input feature.
$g \langle \cdot ;\mathbf{W}^k_{S}\rangle$ means the convolution layer with kernel 7$\times$7.
$\otimes$ is the spatial-wise multiplication operation. Extensive experiments (see~\secref{sec:experiment}) show the effectiveness of these operations in transferring resultful motion patterns from teacher into student branch.

\vspace{-10pt}
\subsection{Explicit Teaching}\label{sec:PD}
\vspace{-3pt}
In addition to the implicit guidance between teacher-student branches, we propose to propagate the motion-guided mask $\mathcal{Z}^M$ containing rich motion cues to teach the decoding of top-three appearance-dominated features $f_k^A, k \in \{3,4,5\}$.

\mySection{Teacher Partial Decoder.}
Given the two groups of cross-modal and multi-level features fused by the appearance- and motion-dominated features, we propose to efficiently utilise them to generate an accurate prediction map at each modality.
Thus, for the time-cost trade-off, we first introduce the teacher partial decoder to aggregate the motion-dominated feature at the top-three level suggested by~\cite{fan2020pra}.
To reduce the computation redundancy while maintaining the representational capability of candidates, we use receptive field blocks~\cite{fan2021concealed} to get the motion-refined features (\ie, $r^M_k = \mathcal{F}^{M,k}_{RF}(f_k^M)$) before the teacher partial decoder $\mathcal{F}_{PD}^{T}$. It is formulated as:
\begin{equation}
    \mathcal{Z}^M = \mathcal{F}_{PD}^{T} [ r^M_3, r^M_4, r^M_5 ].
\end{equation}
Specifically, it can be formulated as two steps:
\begin{equation}
\begin{aligned}
&\textit{Step-I:}~p^{M}_{k} = r^{M}_{k} \otimes \prod_{i=k+1}^{5} g \langle \delta(r^{M}_{i});\mathbf{W}^k_i \rangle, \\
&\textit{Step-II:}~\mathcal{Z}^M = \mathcal{F}_U [p^{M}_{3}, p^{M}_{4}, p^{M}_{5}].
\end{aligned} 
\end{equation}
We term the \textit{Step-I} as feature broadcasting phase, which can broadcast the strong semantic features to the weakly semantic features.
$\prod$ means the element-wise multiplication operation of multiple inputs iterated by $i$, which is parameterized by learnable weights $\mathbf{W}^k_i$.
$\delta(\cdot)$ is an upsampling function to ensure the shape matching of feature.
In \textit{Step-II}, we get the intermediate motion-guided mask $\mathcal{Z}^M$ via typical UNet-shaped~\cite{Ronneberger2015UNetCN} decoder $\mathcal{F}_U$ (removing bottom-two layers).

\mySection{Motion-guided Mask Propagation.}
To effectively leverage the motion-guided mask, we explicitly propagate the motion-guided mask $\mathcal{Z}^M$ to the top-three appearance-dominated features $f_k^{tm}$ derived from the student branch (see~\eqnref{eq:f^tm}). This explicit teaching operation can be formulated as:
\begin{equation}
    f^{et}_k = f^{tm}_k \oplus (f^{tm}_k \otimes \mathcal{Z}^M),
\end{equation}
where $k \in \{ 3,4,5\}$. $\oplus$ and $\otimes$ denote the element-wise addition and multiplication operation, respectively.

\mySection{Student Partial Decoder.}
Adopting the same formulation of teacher partial decoder defined above, we propagate the motion-guided mask $\mathcal{Z}^M$ into the student partial decoder and produce the result $\mathcal{Z}^A$. This process can be defined as:
\begin{equation}
    \mathcal{Z}^A = \mathcal{F}_{PD}^{S}[f^{et}_3, f^{et}_4, f^{et}_5].
\end{equation}
In the inference phase, we serve the output $\mathcal{Z}^A$ followed by \textit{sigmoid} function as the final prediction.

\begin{table*}[tp!]
  \centering
  \scriptsize
  \renewcommand{\arraystretch}{1.0}
  \setlength\tabcolsep{13.2pt}
  \caption{
  Performance comparison with 14 cutting-edge VSOD models on 3 datasets in terms of $\mathcal{M}$, $F_{\beta}$, and $\mathcal{S}$ metrics. 
  The best results in each row are highlighted in \textbf{bold}.
  %
  %
  $\dag$ denote Image salient object detection methods.
  `+M' and `+A' denote the variants that we only train the teacher and student branches, respectively.
  }
  \vspace{-5pt}
  \label{tab:result_comparison}
    \begin{tabular}{l|c||ccc | ccc | ccc}
    \toprule
    & & \multicolumn{3}{c|}{ViSal~\cite{Wang2015ConsistentVS}} &\multicolumn{3}{c|}{DAVIS$_{16}$~\cite{Perazzi2016ABD}} &\multicolumn{3}{c}{DAVSOD$_{35}$~\cite{Fan2019ShiftingMA}} \\
    \cline{3-11}
    Baseline & Pub'Year &$\mathcal{M}$ $\downarrow$&$F_{\beta}$$\uparrow$&$\mathcal{S}$$\uparrow$ &$\mathcal{M}$ $\downarrow$&$F_{\beta}$$\uparrow$&$\mathcal{S}$$\uparrow$ &$\mathcal{M}$ $\downarrow$&$F_{\beta}$$\uparrow$&$\mathcal{S}$$\uparrow$ \\
    \hline
    DSS$^\dag$~\cite{Hou2017DeeplySS} & CVPR'17 & 0.024 & 0.917 & 0.925 & 0.059 & 0.720 & 0.791 & 0.112 & 0.545 & 0.630\cr
    BMPM$^\dag$~\cite{Zhang2018ABM} & CVPR'18 & 0.022 & 0.925 & 0.930 & 0.046 & 0.796 & 0.834 & 0.089 & 0.599 & 0.704\cr
    BASNet$^\dag$~\cite{Qin2019BASNetBS} & CVPR'19 & \textbf{0.011} & 0.949 & 0.945 & 0.029 & 0.818 & 0.862 & 0.110 & 0.597 & 0.670\cr
    SIVM~\cite{Rahtu2010SegmentingSO} & ECCV'10   & 0.199  & 0.521  & 0.611  & 0.211  & 0.461  & 0.551  & 0.291  & 0.299  & 0.491\cr
    MSTM~\cite{Tu2016RealTimeSO} & CVPR'16 & 0.091 & 0.681 & 0.744 & 0.166 & 0.437 & 0.588 & 0.210 & 0.341 & 0.529\cr   
    SFLR \cite{Chen2017VideoSD} & TIP'17  & 0.059  & 0.782  & 0.815  & 0.055  & 0.726  & 0.781  & 0.132  & 0.477  & 0.627\cr
    SCOM ~\cite{Chen2018SCOMSC} & TIP'18 & 0.110 & 0.829 & 0.761 & 0.048 & 0.789 & 0.836 & 0.217 & 0.461 & 0.603\cr
    SCNN~\cite{Tang2019WeaklySS} & TESVT'18 & 0.072 & 0.833 & 0.850 & 0.066 & 0.711 & 0.785 & 0.129 & 0.533 & 0.677\cr
    FCNS~\cite{Wang2018VideoSO} & TIP'18 & 0.045 & 0.851 & 0.879 & 0.055 & 0.711 & 0.781 & 0.121 & 0.545 & 0.664\cr
    FGRNE~\cite{Li2018FlowGR} & CVPR'18 & 0.041 & 0.850 & 0.861 & 0.043 & 0.782 & 0.840 & 0.099 & 0.577 & 0.701\cr
    PDBM~\cite{Song2018PyramidDD} & ECCV'18 & 0.022 & 0.916 & 0.929 & 0.028 & 0.850 & 0.880 & 0.107 & 0.585 & 0.699\cr
    SSAV~\cite{Fan2019ShiftingMA} & CVPR'19 & 0.018 & 0.939 & 0.943 & 0.029 & 0.861 & 0.891 & 0.092 & 0.602 & 0.719\cr
    MGAN~\cite{Li2019MotionGA} & ICCV'19 & 0.015 & 0.944 & 0.944 & 0.022 & \textbf{0.897} & 0.911 & 0.114 & 0.627 & 0.737\cr
    TENet~\cite{Ren2020TENetTE} & ECCV'20 & 0.014 & 0.947 & 0.943  & \textbf{0.021} & 0.894 & 0.905 & 0.078 & 0.648 & 0.753\cr
    \hline
    \multirow{3}{*}{\textbf{\ourmodel~(OUR)}} & + M  & 0.052 & 0.825 & 0.797 & 0.049 & 0.845 & 0.827 & 0.126 & 0.574 & 0.541\cr
     & + A & 0.029 & 0.922 & 0.924 & 0.030 & 0.878 & 0.837 & 0.084 & 0.589 & 0.666\cr
     &\textbf{+ M + A} & 0.018 & \textbf{0.950} & \textbf{0.947} & 0.022 & 0.892 & \textbf{0.912} & \textbf{0.074} & \textbf{0.673} & \textbf{0.760}\cr
    \bottomrule
    \end{tabular}
\end{table*}

\section{Experiment and Analysis}
\label{sec:experiment}
\vspace{-5pt}
\mySection{Implementation Details.}
We train our network on GTX TITAN Xp GPU for computing acceleration with PyTorch.
Following the same pipeline illustrated in~\cite{Li2019MotionGA}.
We first train the teacher branch on the optical flow map generated.
Then, we train the student branch on the DUTS~\cite{wang2017learning} dataset.
Final, we train the dual-branch framework on the training set of DAVIS$_{16}$~\cite{Perazzi2016ABD}.
During the training, we adopt Adam optimizer that the learning rate is initially set to 1e-4 and is decayed by 0.1 every 25 epoch.
We resize the input RGB and optical flow map to 352$^2$ for both the training and test. 
Without any post-processing, the inference speed is 28 \textit{fps}, regardless of flow estimation.
The code and result will be made publicly.

\mySection{Quantitative Comparison.}
To demonstrate the effectiveness of the proposed method, we compare the proposed approach including 11 VSOD methods~\cite{Rahtu2010SegmentingSO,Tu2016RealTimeSO,Chen2017VideoSD,Chen2018SCOMSC,Tang2019WeaklySS,Wang2018VideoSO,Li2018FlowGR,Song2018PyramidDD,Fan2019ShiftingMA,Li2019MotionGA,Ren2020TENetTE} and 3 ISOD methods~\cite{Hou2017DeeplySS,Zhang2018ABM,Qin2019BASNetBS}.
The results in~\tabref{tab:result_comparison} shows the superiority of our \ourmodel~compared to other cutting-edge baselines without any post-processing.

\begin{figure}[t!]
    \centering
    \begin{overpic}[width=\linewidth]{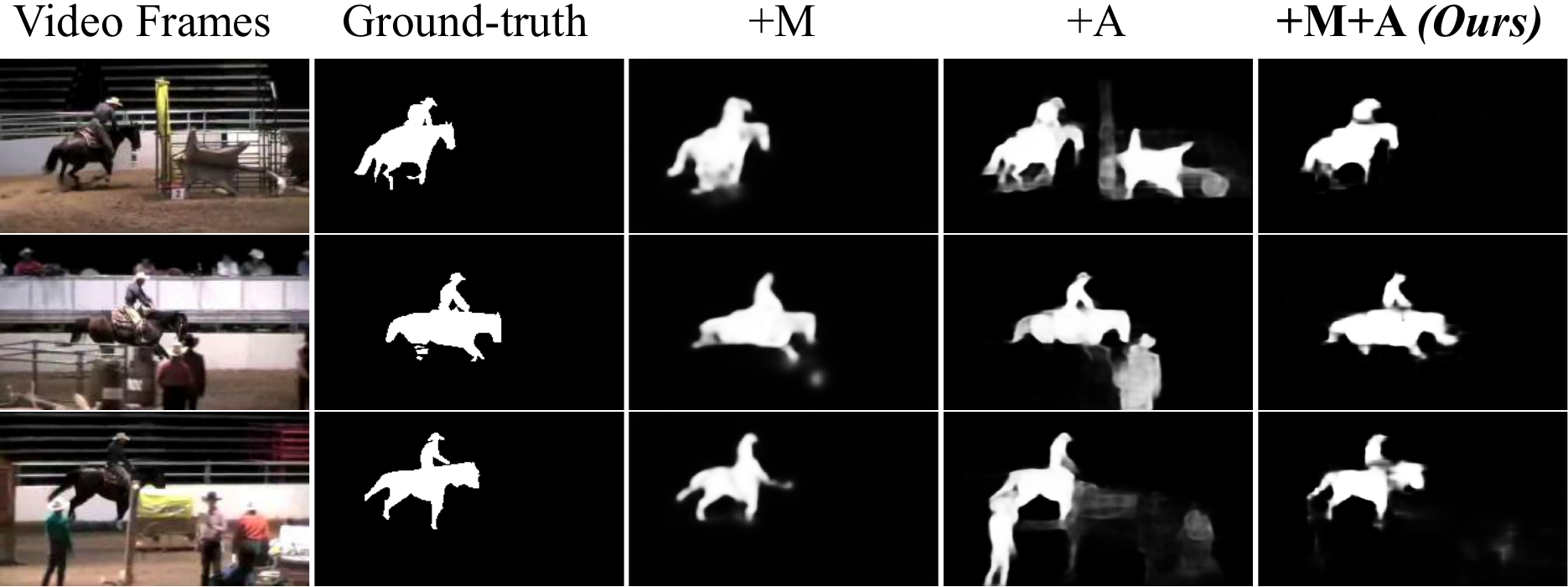}
    \end{overpic}
    \caption{Visual comparison selected from DAVSOD$_{35}$~\cite{Fan2019ShiftingMA}.}
\label{fig:compare_all_datasets}
\end{figure}

\mySection{Qualitative Comparison.}
Following the same train settings in~\tabref{tab:result_comparison}, we compare three variants of \ourmodel, including `+M', `+A', and `+M+A (Ours)'.
As shown in~\figref{fig:compare_all_datasets}, we can observe that our `+M+A' perform satisfactorily in the visual perception compared to other two variants: (a) `+M' only capture coarse location of dynamic objects with fuzzy boundary, and (b) `+A' predict both static and dynamic salient object in the unconstrained scenarios.
Due to the limited space, we refer readers to the supplementary material for more details about the learning strategy, dataset, evaluation metric, visual comparison, and generalizability of the framework.

\mySection{Effectiveness of implicit guidance.}
As shown as in~\tabref{tab:explicit_teaching1}, we conduct ablation studies via decoupling the temporal modulator (TM) to validate the effectiveness of our implicit guidance strategy. 
We remove `CA' and `SA' (No.\#1) and find it is worse than our method (last row), decreasing by 1.6\% in terms of $\mathcal{S}$ on ViSal~\cite{Wang2015ConsistentVS} and 0.9\% in terms of $F_{\beta}$ on DAVIS$_{16}$~\cite{Perazzi2016ABD}.
Besides, we observe that the variants of removing `CA' (No.\#2) or `SA' (No.\#3) can improve the performance of No.\#1, and combine them together (No.\textbf{OUR}) can further boost the performance.

\mySection{Effectiveness of explicit teaching.}
As shown in~\tabref{tab:explicit_teaching}, we further analyse the effectiveness of the explicit teaching strategy via removing the teacher partial decoder (T-PD), student partial decoder (S-PD), and motion-guided mask propagation (`teaching').
We first remove the T-PD on the motion branch (see No.\#4) to validate its effectiveness, the performance degradation on the DAVIS$_{16}$ by a large margin (1.3\% of $\mathcal{S}$).
Then, we verify the necessity of S-PD via decoupling it (see No.\#5) and observe that the variant without it decrease by 0.8\% $F_\beta$ on ViSal.
Further, we remove the motion-guided mask propagation (\ie, yellow lines in~\figref{fig:framework}) before the S-PD (see No.\#6) to verify the effectiveness of the explicit teaching strategy proposed. Comparing to No.\#6, we find that No.\textbf{OUR} with the teaching mechanism improves 0.7\% $\mathcal{S}$ and 0.6\% $F_{\beta}$ in ViSal~\cite{Wang2015ConsistentVS} datasets, which demonstrates that the teaching mechanism is critical to the performance.

\begin{table}[tp!]  
  \centering  
  \scriptsize
  \renewcommand{\arraystretch}{1.1}
  \setlength\tabcolsep{5pt}
  \caption{Ablation studies for implicit guidance. `DB' = Dual-Branch. `CA' = Channel Attention. `SA' = Spatial Attention.}
  \vspace{-10pt}
  \label{tab:explicit_teaching1}  
    \begin{tabular}{l || ccc || ccc | ccc}
    \toprule
     &&&& \multicolumn{3}{c|}{ViSal~\cite{Wang2015ConsistentVS}}&\multicolumn{3}{c}{DAVIS$_{16}$~\cite{Perazzi2016ABD}} \\
    \cline{5-10}
    No. & DB & CA & SA
    & $\mathcal{M}$ $\downarrow$ &$F_{\beta}$ $\uparrow$ &$\mathcal{S}$ $\uparrow$ & $\mathcal{M}$ $\downarrow$ &$F_{\beta}$ $\uparrow$&$\mathcal{S}$$\uparrow$ \\
    \hline
    
    \#1 &\Checkmark &&& 0.022 & 0.935 & 0.934  & 0.029  & 0.883  & 0.879 \\
    \#2 &\Checkmark &\Checkmark & & 0.018 & 0.940 & 0.943 & 0.025 & 0.888 & 0.886 \\
    \#3 &\Checkmark & & \Checkmark & 0.019 & 0.942 & 0.944 & 0.023 & 0.890 & 0.890 \\
    \textbf{OUR} &\Checkmark&\Checkmark & \Checkmark& \textbf{0.018} & \textbf{0.947} & \textbf{0.950} & \textbf{0.022} & \textbf{0.892} & \textbf{0.912} \\
    \bottomrule  
    \end{tabular} 
\end{table}

\begin{table}[tp!]  
  \centering  
  \scriptsize
  \renewcommand{\arraystretch}{1.3}
  \setlength\tabcolsep{3.5pt}
  \caption{Ablation studies for explicit teaching strategy.}
  \vspace{-10pt}
  \label{tab:explicit_teaching}
    \begin{tabular}{l||ccc||ccc|ccc}  
    \toprule
     &&&& \multicolumn{3}{c|}{ViSal~\cite{Wang2015ConsistentVS}}&\multicolumn{3}{c}{DAVIS$_{16}$~\cite{Perazzi2016ABD}} \\
    \cline{5-10}
    No. & T-PD & S-PD & Teaching & $\mathcal{M} \downarrow$ &$F_{\beta}  \uparrow$ &$\mathcal{S} \uparrow$ & $\mathcal{M} \downarrow$ &$F_{\beta}  \uparrow$ &$\mathcal{S}  \uparrow$ \\
    \hline
    \#4 & &\Checkmark & \Checkmark& 0.022 & 0.942 & 0.939 & 0.026 & 0.883 & 0.889 \\
    \#5 &\Checkmark & & \Checkmark & 0.020 & 0.939 & 0.933 & 0.028 & 0.887 & 0.883 \\
    \#6 &\Checkmark & \Checkmark & & 0.021 & 0.941 & 0.943 & 0.025 & 0.890  & 0.890 \\
    \textbf{OUR} &\Checkmark &\Checkmark& \Checkmark &\textbf{0.018} & \textbf{0.947} & \textbf{0.950} & \textbf{0.022} & \textbf{0.892} & \textbf{0.912} \\
    \bottomrule  
    \end{tabular} 
\end{table}

\vspace{-5pt}
\section{CONCLUSION}
\vspace{-5pt}
In this paper, we emphasis the importance of both \textit{implicit guidance and explicit teaching} strategies for spatial-temporal representations.
Our dual branch architecture is attributed to two key designs: 
\textit{(i)} adopting temporal modulator implicitly transmits the representative motion-dominated cues into an appearance-dominated branch; and 
\textit{(ii)} using the motion-guided mask to explicitly teach the feature aggregation of appearance-dominated branch.
Extensive experiments on three datasets demonstrate that the proposed \ourmodel~performs competitively compared to 14 cutting-edge approaches.
%




\newpage

\bibliographystyle{IEEEbib}
\bibliography{refs}

\end{document}